\documentclass[letterpaper, 10 pt, conference]{ieeeconf}  

\IEEEoverridecommandlockouts                              

\usepackage[utf8]{inputenc} 
\usepackage[T1]{fontenc}    
\usepackage{hyperref}       
\usepackage{url}            
\usepackage{booktabs}       
\usepackage{amsfonts}       
\usepackage{nicefrac}       
\usepackage{microtype}      
\usepackage{xcolor}         
\usepackage{graphicx} 
\usepackage{amsmath} 
\usepackage{graphicx}
\usepackage{booktabs}
\usepackage[accsupp]{axessibility}  
\usepackage{amsmath}
\usepackage{amssymb}
\usepackage{booktabs}
\usepackage{pifont}
\usepackage{algorithm}
\usepackage{algpseudocode}
\usepackage{times}
\usepackage{epsfig}
\usepackage{graphicx}
\usepackage{amssymb}
\usepackage{float}
\usepackage{booktabs}
\usepackage{subcaption}
\usepackage{multirow}
\usepackage{xcolor}
\mathchardef\mhyphen="2D
\usepackage[utf8]{inputenc}
\usepackage{comment}
\usepackage{tabularx,colortbl}
\usepackage{xparse,mathtools}
\usepackage{diagbox}
\usepackage{adjustbox,lipsum}
\definecolor{Gray}{gray}{0.93}
\usepackage{breqn}
\usepackage{makecell}

\usepackage{tabulary,overpic}

\newlength\savewidth

\newcommand{\bi}{\mathbf{i}}

\newcommand{\bo}{\mathbf{o}}

\newcommand{\bV}{\mathbf{V}}
\newcommand{\bw}{\mathbf{w}}

\newcommand{\cO}{\mathcal{O}}

\newcommand{\cW}{\mathcal{W}}

\makeatletter
\DeclareRobustCommand\onedot{\futurelet\@let@token\@onedot}
\def\@onedot{\ifx\@let@token.\else.\null\fi\xspace}
\def\eg{e.g\onedot} 
\def\ie{i.e\onedot} 
\def\cf{cf\onedot} 
\def\etc{etc\onedot}

\def\etal{et~al\onedot}

\makeatother

\newcommand{\boldparagraph}[1]{\vspace{0.2cm}\noindent{\bf #1:}}

\definecolor{darkgreen}{rgb}{0,0.7,0}

\colorlet{lr}{red!70!white}
\colorlet{lb}{cyan!70!white}
\colorlet{lo}{orange!70!white}
\colorlet{lm}{magenta!70!white}

\definecolor{lb}{HTML}{FFEFCC}
\definecolor{lo}{HTML}{CCFFEF}
\definecolor{lm}{HTML}{CDCCFF}

\usepackage{cite}

\usepackage{hyperref}

\usepackage{orcidlink}
\usepackage[capitalize]{cleveref}
\Crefname{section}{Sec.}{Secs.}
\Crefname{section}{Section}{Sections}
\Crefname{table}{Table}{Tables}
\Crefname{table}{Tab.}{Tabs.}

\usepackage{wrapfig}
\usepackage{cancel}
\usepackage[normalem]{ulem}
\usepackage[utf8]{inputenc}
\usepackage{pifont}
\usepackage{newunicodechar}
\newunicodechar{✓}{\ding{51}}
\newunicodechar{✗}{\ding{55}}

\newcommand{\green}{\textcolor[rgb]{0.1,0.6,0.1}}
\newcommand{\darkred}{\textcolor[rgb]{0.6,0.1,0.1}}
\definecolor{lightgray}{rgb}{0.89, 0.89, 0.89}

\definecolor{i2}{HTML}{DEFFCE}
\definecolor{i1}{HTML}{98FB98}

\definecolor{v2}{HTML}{D3E0FF}
\definecolor{v1}{HTML}{89CFF0}

\usepackage[most]{tcolorbox}
\usepackage[dvipsnames]{xcolor}

\usepackage{wrapfig}
\usepackage[framemethod=TikZ]{mdframed}
\newcounter{theo}[section]

\usetikzlibrary{calc,shadows.blur}
\tcbuselibrary{skins}
\newtcolorbox{resp}[1][]{%
  enhanced jigsaw,
  colback=black,%
  colframe=topBlue,
  size=small,
  boxrule=1pt,
  title=\textbf{\textit{Answer}},
  halign title=flush center,
  coltitle=black,
  breakable,
  drop shadow=black!20!white,
  attach boxed title to top left={xshift=0.5cm,yshift=-\tcboxedtitleheight/2,yshifttext=-\tcboxedtitleheight/2},
  minipage boxed title=3cm,
  boxed title style={%
    colback=white,
    size=fbox,
    boxrule=1pt,
    boxsep=2pt,
    underlay={%
      \coordinate (dotA) at ($(interior.west) + (-0.5pt,0)$);
      \coordinate (dotB) at ($(interior.east) + (0.5pt,0)$);
      \begin{scope}
        \clip (interior.north west) rectangle ([xshift=3ex]interior.east);
        \filldraw [white, blur shadow={shadow opacity=0, shadow yshift=-.75ex}, rounded corners=2pt] (interior.north west) rectangle (interior.south east);
      \end{scope}
      \begin{scope}[white!80!white]
       
      \end{scope}
    },
  },
  #1,
}

\colorlet{LightLavender}{Lavender!40!}
\tcbset{on line, 
        boxsep=4pt, left=0pt,right=0pt,top=0pt,bottom=0pt,
        colframe=white,colback=LightLavender,  
        highlight math style={enhanced}
        }

\newtcolorbox{mybox}[3][]
{
  colframe = #2!25,
  colback  = #2!7,  
  coltitle = #2!80!black,  
  #1,
}

\title{\LARGE \bf Time-Aware Assistive Navigation}

\author{
Masaki Kuribayashi$^{*1,2}$, Zhongkai Shangguan$^{*2}$, and Eshed Ohn-Bar$^{2}$
\thanks{----------------}
\thanks{$^{*}$ indicates equal contribution. $^{1}$Waseda University; $^{2}$Boston University.}
}

\begin{document}

\def\eg{{\it e.g.}}
\def\cf{{\it c.f.}}
\def\ie{{\it i.e.}}
\def\etal{{\it et al. }}
\def\etc{{\it etc}}

\maketitle

\begin{abstract} 
Can interactive vision-and-language agents learn not just what to say but also \textbf{\textit{when}} to say it? 
Current language models rarely plan over whether and when to realize a real-time response to a user. However, providing accurate and timely support for human decision-making, such as when guiding visually impaired individuals through urban environments, requires careful real-time responsiveness--poorly timed responses can distract users or add unnecessary cognitive load.
As a machine intelligence challenge for Multimodal Large Language Model (MLLM)-based agents, we introduce a large-scale multimodal benchmark for an egocentric, assistive navigation task in complex outdoor environments. Using this benchmark, we uncover a fundamental limitation of off-the-shelf MLLMs in delivering safe and time-sensitive navigation instructions, even with model fine-tuning on substantial amounts of data. 
We then demonstrate that a simple yet effective modification of the model, including direct supervision to predict the underlying reason for each instruction, yields significant performance gains across open-loop, closed-loop, and sim-to-real generalization settings. However, our analysis highlights persistent challenges in temporal reasoning, safety-critical object awareness, and relational and distance understanding. 
To advance the development of scalable assistive agents, we will release our simulation, benchmark, and code (available at project website: \url{https://timeli-icra.github.io/}). 
\end{abstract}

\section{Introduction}
\label{sec:introduction}

Effective assistive vision-and-language models need to go beyond merely describing their surroundings--—they must understand users' needs, anticipate risks, and determine when to intervene, or to remain silent.
To realize their potential as real-world assistive agents~\cite{gurari2018vizwiz,fei2022searching, puig2020watch, li2023behavior,assister2022huang}, such as wearable devices and robotic platforms guiding visually impaired users through unfamiliar environments and chaotic city streets, they must be designed to provide proactive, safe, and timely support. 
Do current Multimodal Large Language Models (MLLMs), have the necessary spatial, temporal, and planning capabilities needed to maintain continual situational awareness and provide seamless assistance to diverse users?

\begin{figure}[!t]
    \centering
    \includegraphics[width=3.4in, trim=0pt 235pt 100pt 10pt, clip]{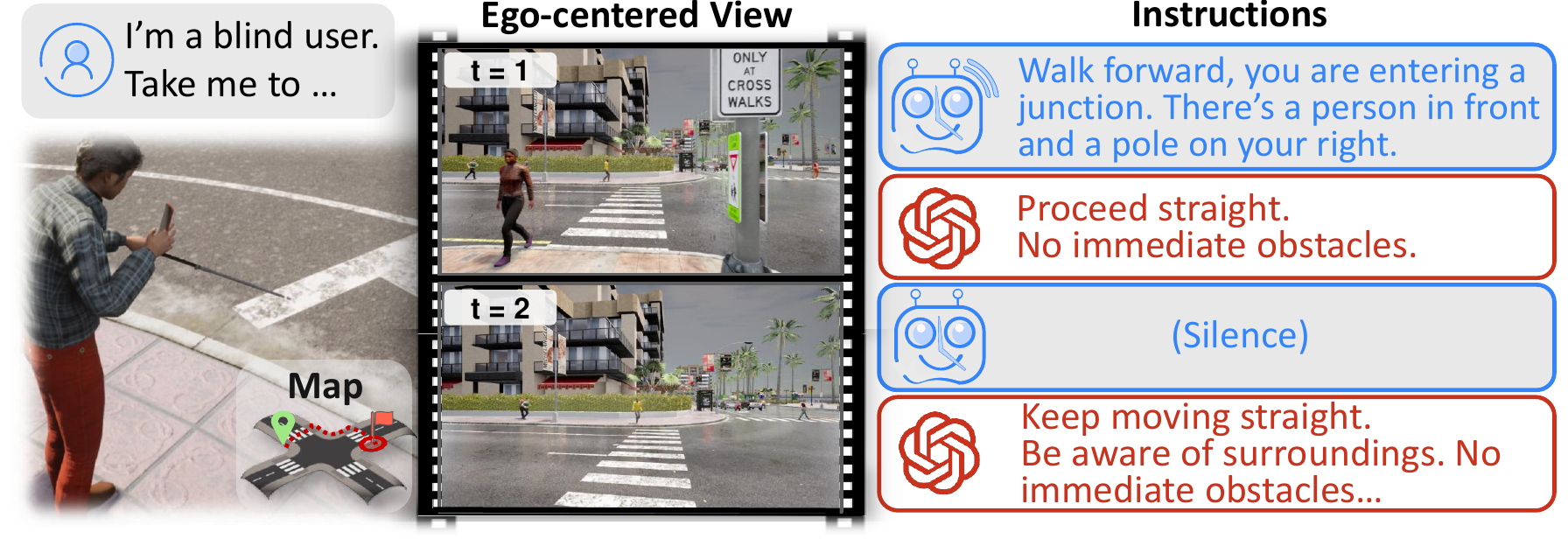} 
    \vspace{-0.3cm}
    \caption{\textbf{Step-by-Step Video Instruction Generation Task.} 
  We introduce \textbf{TIMELI}, a novel machine intelligence task and model for providing temporally-aware, concise, and safe assistance to blind users navigating toward a goal. 
  Based on our analysis, we demonstrate how current MLLMs, \eg, ChatGPT (depicted in the figure), provide ineffective and overly verbose instructions for our task. In contrast, the TIMELI-based agent reasons about both what to say and when to say it, ensuring alignment with the needs of blind users, \eg, remaining silent at intersections to reduce cognitive load and prevent hazardous situations, which is a standard strategy employed by mobility guides.
    }
      \label{fig:teaser_ver2}
     \vspace{-0.50cm}
\end{figure}

Consider the recent Be My Eyes and OpenAI~\cite{bemyeyes2} demonstration of GPT-4o~\cite{gpt4o} as an interactive, real-time outdoor assistant for a blind user navigating urban environments (we show overall task example in Fig.~\ref{fig:teaser_ver2}). 
While promising in principle, even minor errors in the content and timing of an instruction in this context could have serious consequences, such as leading the user off course or into a dangerous situation. For example, the user might be misled when the model mixes up a single yet crucial word in its response, \eg, \textit{``{left}''} with \textit{``{right},''} which is a common error based on our planning-oriented analysis. Eventually realizing the error, our user may attempt to cross safely while listening to surrounding signals and traffic. However, the verbose model may continue to provide lengthy instructions (another common behavior~\cite{saito2023verbosity,zhao2024long}) \textit{while the user is crossing}, causing significant distraction and cognitive load~\cite{crossblind,Lee2020Emerging}. In this scenario, the user could attempt to prompt the model with more examples and specifications regarding instructions, \eg, \textit{``The intersection was on the left, not on the right, and don't talk at intersections when user is crossing, it's dangerous.''} However, the model does not follow the request as it was not trained with the objective of acting as a mobility guide for a blind user~\cite{ouyang2022training,mack2024they,gallegos2024bias}, with very few of the training samples, if any, included blind individuals (\eg, none in Webvid-2M~\cite{bain2021frozen}, VALOR~\cite{chen2023valor}). In this work, we develop a suitable benchmark and model for instilling usability knowledge into MLLM-based agents. 

Despite recent progress in temporal video understanding, surprisingly little work has explored safe, real-time assistive navigation instruction. Instead, most approaches focus on video captioning and generic summarization, \ie, producing broad, task-agnostic descriptions~\cite{shen2024longvu, vaishnavi2024video, luo2023valley, ren2024timechat, song2024moviechat,zhang2023video,lin2023video,yang2022tubedetr,gurari2018vizwiz} rather than goal-oriented, real-time assistance. For example, VizWiz~\cite{gurari2018vizwiz}, widely used in MLLM studies~\cite{lin2024vila}, primarily focuses on single-frame tasks, such as close-up object and text recognition. Similarly, prior work in Vision-and-Language Navigation (VLN)~\cite{wang2023learning, puig2020watch, zhou2024navgpt, wang2023lana, zeng2023kefa, fan2025navigation, assister2022huang, kong2024controllable, wang2022less} generally studies static and simple environments, \ie, without fine-grained relational understanding, reactive dynamic obstacles, complex outdoor layouts, weathers, or continual assistance for disabled users. Indeed, even with fine-tuning on ample data, our analysis reveals that most off-the-shelf models demonstrate poor performance across key task metrics.

\boldparagraph{Contribution} 
{
Our goal is to advance user-aware, dynamic, and safe assistive agents. First, we introduce \textbf{TIMELI}, an extensive video-based, TIME-aware Language Instruction benchmark designed to accurately evaluate safety-critical temporal reasoning in multimodal AI systems for blind users. 
TIMELI includes both synthetic and real-world data, and was developed in collaboration with mobility guides and blind users to model expert-level guidance with context-aware instructions, including directing users to walk alongside buildings, avoid cane-length obstacles, and remain silent at intersections to ensure instructions are situationally appropriate, safe, and timely. Second, we evaluate several models on TIMELI and demonstrate their failure in both open-loop (video-based) and closed-loop (interactive) settings, even after fine-tuning on synthetic data to reduce the domain gap.}

\section{Related Work}
\label{sec:related}
\boldparagraph{Video and Temporal Understanding} 
Our assistive navigation task requires precise modeling of spatio-temporal relationships in video data. Recently, image encoders such as CLIP~\cite{radford2021learning} and ViT~\cite{vit} have been adapted to encode video data~\cite{yang2022tubedetr,zhang2023video,vaishnavi2024video,khattak2024complex,cai2024temporalbench,wu2024longvideobench,li2024aria,zeng2024matters,wang2024qwen2}, enabling advancements in tasks such as video summarization~\cite{lin2023video,shen2024longvu,luo2023valley,wang2024qwen2,lin2024vila,chen2023videollm,yang2023vid2seq} and turn-by-turn question answering~\cite{ren2024timechat,maaz2023video,li2023videochat,song2024moviechat}. However, visual understanding capabilities of state-of-the-art open-source and closed-source models~\cite{gpt4o,touvron2023llama} remain limited~\cite{tong2024eyes,cai2024temporalbench,kaniwa2024chitchatguide}, raising concerns about their usability as interactive contexts. Moreover, prior benchmarks for question answering, \eg, VizWiz~\cite{gurari2019vizwizpriv,bigham2010vizwiz,tseng2022vizwiz} (often used in MLLM evaluations~\cite{lin2024vila}), do not generally consider the issue of planning for \textit{when to cue} for a response, yet both how and when to realize a cue can be tightly integrated in our context. Similarly, while recent work in generalized time-aware instruction tuning~\cite{ren2024timechat} provides an initial step in reasoning over temporal observations, current models cannot capture interactions between fine-grained landmarks, 3D spatial planning, and safety and conciseness constraints within dynamic, real-time contexts. Our analysis reveals that existing models do not generalize to such instruction tasks.

\boldparagraph{User-Aware Navigation Instruction Generation} In contrast to the standard Vision-and-Language Navigation (VLN) task~\cite{ku2020room,hong2020sub,qi2020object,liu2023bird,zhang2024vision,moudgil2021soat,xu2024flame,gu2022vision,wu2023vision,zhou2024navgpt,zhang2024navid,ping,anderson2018vision,anderson2021sim,chen2019touchdown}, which involves interpreting instructions into low-level navigation actions, we are concerned with instruction generation to a user. Within our task, most prior work considers static (\eg, on R2R~\cite{anderson2018vision}) and simplified settings~\cite{daniele2017navigational,fried2018speaker,wang2023learning,zhao2021evaluation,zeng2023kefa,fan2025navigation,assister2022huang,kong2024controllable,wang2022less}, without video reasoning, prior interaction history, instructional timing, or real-time user state. Similarly, step-by-step commercially-available services, \eg, based on Google Maps, are limited as they tend to generate instructions based on various planning and location heuristics~\cite{fogli2020universal,sato2019navcog3}, thus not considering issues with GPS error, user preferences, or accessibility needs. To re-iterate, prior approaches and models lack the ability to determine optimal timing for instructions, even though timing and content are inherently linked. Instead, an instruction may always be sampled from such models for any instantaneous state, even in cases where it may not be needed or could be dangerous, \eg, during crossings~\cite{crossblind,Lee2020Emerging}. In this work, our task requires the model to sequentially decide what to say and when, which involves an intricate planning task over the high-level plan, environmental hazards, and overall task and path adherence.

\begin{table}[!t]
\centering
\caption{\textbf{Comparing Vision-and-Language Navigation Datasets.} Compared to prior datasets, our benchmark includes rich contextual instructions and temporal information, specifying when and what to say to guide a visually impaired user. Additionally, we emphasize the involvement of dynamic obstacles (\eg, pedestrians) in outdoor navigation scenarios. 
For comparison with other non-temporal benchmarks, we also note the size of each dataset (total number of images in TIMELI is $1{,}233{,}616$). While the synthetic benchmark is procedurally generated to incorporate diverse factors (e.g., weather, time of day, safety hazard), we further curate a real-world dataset sourced from an in-situ study and manually annotated YouTube videos, reserved for validation.
}

\label{tab:dataset_comparison}
\resizebox{\columnwidth}{!}{%
\begin{tabular}{lcccccc}
\toprule
    Dataset & 
    Dynamic & 
    Context & 
    \multicolumn{1}{c}{\begin{tabular}[c]{@{}c@{}}Temporal\\ Information\end{tabular}} &
    \multicolumn{1}{c}{\begin{tabular}[c]{@{}c@{}}For Blind\\ End-User\end{tabular}} &
    Size & 
    Collection \\ \midrule
R2R~\cite{anderson2018vision}           & \darkred{✗}  & Indoor  & \darkred{✗}  & \darkred{✗}  & 21{,}567 Images  & Real-World \\
CVDN~\cite{thomason2020vision}          & \darkred{✗}  & Indoor  & \darkred{✗}  & \darkred{✗}  & 7{,}000 Images & Real-World \\
REVERIE~\cite{qi2020reverie}       & \darkred{✗}  & Indoor  & \darkred{✗}  & \darkred{✗}  & 21{,}702 Images & Real-World \\
Touchdown~\cite{chen2019touchdown}     & \green{✓} & Outdoor & \darkred{✗}  & \darkred{✗}  & 9326 Images & Real-World \\
Talk the Walk~\cite{de2018talk} & \green{✓} & Outdoor & \darkred{✗}  & \darkred{✗}  & 10{,}000 Images & Real-World \\
RxR~\cite{ku2020room}           & \darkred{✗}  & Indoor  & \darkred{✗}  & \darkred{✗}  & 126{,}069 Images & Real-World \\
WAY~\cite{hahn2020you}           & \darkred{✗}  & Indoor  & \darkred{✗}  & \darkred{✗}  & 6{,}154 Images   & Real-World \\
UrbanWalk~\cite{assister2022huang} & \green{✓} & Outdoor & \darkred{✗}  & \green{✓} & 399{,}126 Images & Synthetic \\
 \midrule 
 \rowcolor{lightgray}
\textbf{TIMELI - Sim } & \green{✓} & Outdoor & \green{✓} & \green{✓} & 67{,}101 Videos  &   Synthetic \\ 
 \rowcolor{lightgray}
\textbf{TIMELI - Real} & \green{✓} & Outdoor & \green{✓} & \green{✓} & 2{,}000 Videos  &  Real-World \\
\bottomrule
\end{tabular}%
}
\vspace{-0.55cm}
\end{table}


\boldparagraph{Designing Assistive Systems}
Our ultimate goal is to develop a system that optimally assists users by dynamically generating navigation guidance instructions. Notably, the design of accessible navigation systems remains an ongoing area of research~\cite{hoogsteen2022beyond,muller2022traveling,kim2024text}. Moreover, appropriate timing of instructions is as crucial as their content to avoid overwhelming the user’s cognitive load, especially given that the required information can change drastically depending on time and situation in diverse real-world environments~\cite{martinez2014cognitive,Lee2020Emerging,lee2022opportunities,guerreiro2018context,ahmetovic2019impact,ohn2018modeling,envfactors}. While various practical systems have been developed, using modalities like vibration~\cite{kuribayashi2021linechaser,martinez2014cognitive,katzschmann2018safe} and sonification~\cite{presti2019watchout,ahmetovic2019sonification,yoon2019leveraging}, audio-based language remains a commonly preferred approach by users~\cite{arditi2013user,li2016isana,sato2019navcog3,kayukawa2022HowUsers}. 
However, these audio cues are typically manually crafted, limiting adaptability to new environments or situations.
Remote sighted human assistance~\cite{ohn2025navigating,Kamikubo2020Support,Lee2020Emerging,lee2022opportunities,chang2024worldscribe}, in which a sighted assistant conveys instructions remotely, has gained adoption and offers flexible guidance, but its usage is costly.
Several preliminary and small-scale approaches based on MLLMs have been recently developed~\cite{wang2024visiongpt,yang2024viassist,song2024guide,kaniwa2024chitchatguide}, while relying on rule-based timing for instructions~\cite{wang2024visiongpt} or focus on basic captions rather than active goal-driven navigation~\cite{kaniwa2024chitchatguide,chang2024worldscribe,yang2024viassist,zhang2024enhancing}. We referred to user responses and failure scenarios in these prior studies, as well as semi-structured interviews with guides and blind users when developing our novel simulation and benchmark. A key finding is that effective instruction can be nuanced and contextual, even around basic turns, \eg, \textit{``find the building, you are inside a recessed doorway, step back, now turn left, trace the building, careful there's an overhead branch.''} Another common principle is to remain silent at intersections, allowing the navigator to rely on mobility skills learned through orientation and mobility training. Here, unnecessary instructions could pose a danger~\cite{Lee2020Emerging} (our benchmark directly quantifies performance in this task). Finally, users prefer avoiding excessive amounts of information (a known issue in LLMs~\cite{saito2023verbosity,zhao2024long}) and instead prioritizing certain information at the right time, \eg, obstacles in cane length~\cite{hoogsteen2022beyond}. Thus, while focusing on accessibility, our study provides a benchmark for social machine intelligence and a step toward generalized assistive agents that can holistically understand user needs.

\begin{figure*}[!t]
    \centering
    \includegraphics[width=0.9\textwidth, trim=0pt 420pt 180pt 10pt, clip]{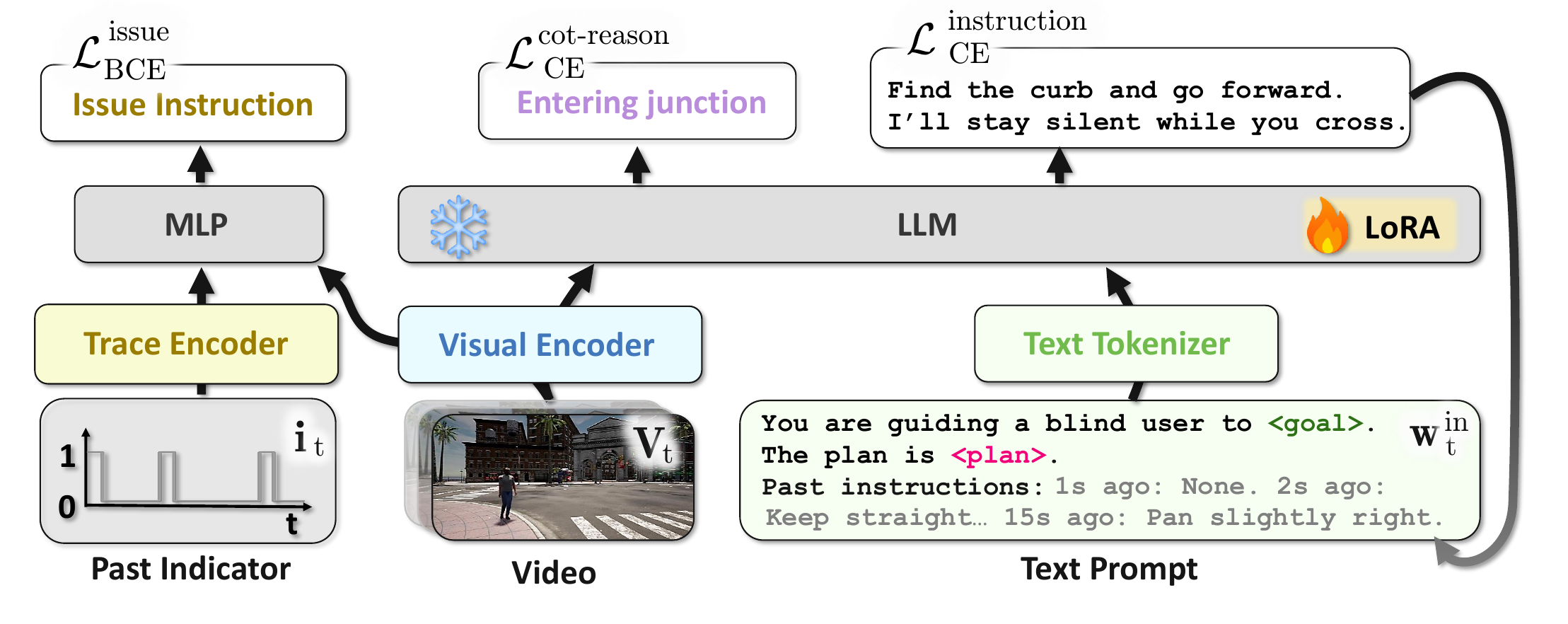}
    \vspace{-0.3cm}
\caption{\textbf{Model Structure and Task Overview.} The model takes input video, plan (defined from a set of map-based waypoints), and task specifications with historical instructions (Sec.~\ref{sec:formal}). We explicitly supervise for instructional reason (Sec~\ref{sec:model}), which we find improves model guidance performance. 
    }
  \label{fig:model}
  \vspace{-0.3cm}
\end{figure*}

\vspace{-0.3cm}
\section{Method}
\label{sec:method}
The goal of our benchmark is to evaluate how well MLLMs can generate time-sensitive and situationally appropriate navigation instructions for blind users, with an overview of model components shown in Fig.~\ref{fig:model}.
In our task, we assume a scenario where a blind person navigates using an assistive system in an outdoor setting, equipped either on a smartphone~\cite{sato2019navcog3,kacorri2016supporting} or a wearable device~\cite{li2016isana}, with an RGB camera to perceive the front field of view. 
The system sequentially observes the front view and issues guidance instructions to the user. 
Following common systems that utilize localization information (\eg, GPS) and destination information, we also assume access to a noisy location and a high-level path planner to the destination~\cite{Dosovitskiy17CARLA}. We formalize our task in Sec.~\ref{sec:formal}, followed by a description of our benchmark creation (Sec.~\ref{sec:benchmark}) and network optimization process (Sec.~\ref{sec:model}).
\vspace{-0.3cm}
\subsection{Task Formulation}
\label{sec:formal}

Time-aware instruction generation is a sequential decision-making task, where the goal at each time step $t$ is to generate an \textit{output instruction} $\bw^{\text{out}}_t = (w_t^1,\dots,w_t^{N_w}) \in \cW$, consisting 
of $N_w$ word tokens, \ie, $w_t \in [1,D]$, $D$ being the vocabulary size. 
At inference time, we assume access to a stream of \textit{observations} $\bo_t = (\bV_t, \bw_t^{\text{in}}) \in \cO$
corresponding to: (1) an ego-centered \textit{video} $\mathbf{V_t} \in \mathbb{R}^{W \times H \times 3 \times T}$ of resolution $W \times H$, \ie, including the current image and past $T-1$ frames ($T = 1$ for image-based models); and (2) a language-based input to the model $\bw^{\text{in}} \in \cW$, comprising \textit{user and task specifications}, a high-level navigation \textit{plan} toward a goal represented as a command and a sequence of 2D waypoints~\cite{zhang2022selfd,zhang2024feedback}. We note that the high-level plan only provides coarse guidance for the final instructions, as it is based on a static, standard-definition map. We also leverage the \textit{history of prior instructions}, as further detailed below.

\boldparagraph{Task Specification and History}
Our task includes detailed specifications of the user and task:

\definecolor{mintwhite}{RGB}{248,255,242}
  
\begin{center}
\begin{tcolorbox}[enhanced,
    colback=mintwhite,
    colframe=black,
  drop fuzzy shadow,watermark color=white]
  \centering
\begin{center}
\begin{quote} \textit{``You are guiding a blind user. You will need to instruct the user to stay on the path to the goal. Notify them of immediate turns and obstacles within 1.5m to avoid. Keep instructions at junctions minimal, for example...''}
\end{quote}
\end{center}
\end{tcolorbox}
\end{center}

The task specification also includes additional relevant context, specifically the goal: \textit{``The blind person needs to approach a relative goal: [x,y] = $[x_{dest}, y_{dest}]$,''} where $[x_{dest}, y_{dest}] \in \mathbb{R}^2$ is the next waypoint. Additionally, a text-based command is used to specify the high-level navigation plan, derived from the next goal waypoint, \eg, \emph{\text{``turn left''}}. The plan (out of seven possibilities) is computed based on the angle between the goal coordinate and the agent’s heading direction of the user. Finally, the task specification input also incorporates a history of prior instructions over a time window of length T, along with their associated timestamps, \eg, \textit{``2 seconds ago, the instruction was: pedestrian on your left. 5 seconds ago, the instruction was: go forward...''} Crucially, the task specification includes an explicit instruction requiring the model to output \textit{``None''} when no user feedback is necessary or contextually appropriate. This ensures that models learn to remain silent when redundant or distracting guidance would be unhelpful.

\vspace{-0.1cm}
\subsection{The TIMELI Benchmark}
\label{sec:benchmark}
\vspace{-0.1cm}

We introduce \textit{TIMELI}, the first video-based time-aware benchmark designed for navigating pedestrians in dynamic urban environments. Despite a large number of vision-and-language navigation benchmarks, there is currently no \textit{video-based} dataset or environment for training and evaluating models that deliver timely and safety-aware navigation instructions to humans. Moreover, in interactive domains where collection and annotation of real-world human data can be difficult, realistic synthetic environments offer a scalable, privacy-preserving, and reproducible solution. 

\boldparagraph{Design Through Collaboration with Mobility Guides} 
In conjunction with guidance from existing literature, we conducted a pilot study with ten blind participants and local Orientation and Mobility guides to explore remote navigation guidance. Participants wore chest-mounted cameras~\cite{Kamikubo2020Support}, while guides gave directions via Zoom using Google Maps. With IRB approval, we recorded video and audio for analysis.
From observations and feedback, we found that frequent, brief cues help maintain orientation, especially in complex environments. Guides confirmed obstacle-free paths, referenced nearby walls for alignment, and used simple directions like \textit{``left''}, or \textit{``right''} or clock-face terms. Numeric distances were avoided in favor of natural phrases like ``keep walking straight.'' These insights informed the TIMELI benchmark’s instruction timing and phrasing.

\boldparagraph{Multimodal Temporal Data Collection} 
As real-world data collection is inherently constrained by safety concerns, data scarcity, and privacy issues, synthetic data has emerged as a promising solution, enabling the creation of artificial data that mimics real-world patterns~\cite{hu2021sail,fabbri2021motsynth,cai2024playing}.
We leverage the open-source simulation environment based on CARLA~\cite{zhang2021x,Dosovitskiy17CARLA} to collect multimodal data for the pedestrian navigation task.
While CARLA is mainly used for autonomous driving, we adapt it to simulate first-person pedestrian data under varied conditions. Pedestrians are randomly spawned on sidewalks with random destinations, following A* paths that comply with traffic rules.
As pedestrians move, we capture multimodal data (location, ego-centric images, depth, and semantic segmentation) at 1 FPS. Intersection locations are extracted due to their navigation complexity. Our data spans six weather conditions and five towns. 

\vspace{-0.2cm}
\boldparagraph{Time-Aware Navigation Instruction Generation}
Our procedural navigation instruction generation process considers both the content and the timing of the instructions, ensuring clear and essential communication that supports effective navigation without overwhelming the pedestrian. 
Specifically, the content of instructions encompasses two aspects: directional cues (\eg, \textit{``Now, turn slightly right and continue walking''}) and environmental descriptions (\eg, \textit{``Be careful, there's a pedestrian in front of you''}). 
 {
Directional cues are generated based on the relative displacement between the pedestrian's current position and the position one second earlier. 
For example, the system instructs the pedestrian to walk forward when the change is within 30$^\circ$, uses the term ``slightly'' for turns under 75$^\circ$, and provides explicit left or right turn instructions for direction changes exceeding 75$^\circ$.
}
Repeated alerts for the same obstacle are removed within a specified time frame, except when an obstacle is detected within one meter (cane length). 
Immediate instructions are also issued when a change of direction is necessary. 
In complex scenarios such as intersections, we deliver instructions with warnings for upcoming turns or obstacles prior to entering the intersections. 
Upon entering or exiting intersections, the instruction includes phrases like \textit{``You are entering/crossing the junction''} or \textit{``You have exited the junction.''}
To validate our simulation data’s real-world relevance, we curated an additional test set from YouTube videos. 
 {
We developed a custom annotation interface where annotators review frames in a video by editing a text box that shows alongside suggested instructions generated by GPT-4o. 
The interface displays four consecutive frames, with the current frame on the left and the next three seconds to the right or a GIF (1 FPS) for dynamic visualization. 
}
Human annotators manually matched scenarios to the user study, followed mobility guides’ instructions, and annotated the navigator’s direction.

\boldparagraph{Data Statistics}
Overall, we collected 67K navigation videos, where each navigation video contains $T$ frames.
Out of the 67K samples, 79.1\% of the data contained none (remained silent), and 20.9\% had an instruction issued. 
Among the spoken instructions, 81.0\% directed to move forward, 17.6\% directed to turn left or turn right, and the rest directed to turn around or to stop. 
The average length of the spoken instruction was 11.56 words, with a standard deviation of 4.59 words, with a maximum instruction length of 32 words and a minimum of two words (\textit{``Turn around.''}).
Additionally, 21.6\% of the data were recorded at intersections. 
Examples of the instructions include: ``\textit{Turn slightly right and continue walking,}'' ``\textit{Okay, keep walking straight. Be careful, pedestrian in your front,}'' and ``\textit{Now, turn slightly right and keep walking. Fence on your right.}''

\subsection{Model Architecture} 
\label{sec:model}

Our task extends prior MLLM-based agents to real-time settings. Given the challenges associated with our task, we refine the instruction space and introduce efficient and targeted supervision for chain-of-thought (CoT) reasoning. Moreover, to further enhance the instruction generation in real-time scenarios, we incorporate auxiliary supervision for timing.

\boldparagraph{Reason Prediction}
To improve the model’s decision-making capabilities, we restructure the output format of $\bw^{\text{out}}$ so that the model first predicts the rationale behind an instruction before generating the instruction itself. The reason prediction is explicitly supervised, representing the reasoning behind each guidance instruction and is categorized into one of eight distinct labels:\emph{\text{``remain silent,''}} \emph{\text{``remain silent in junction,''}} \emph{\text{``enter junction,''}} \emph{\text{``exit junction,''}} \emph{\text{``obstacle in front,''}} \emph{\text{``constant instruction,''}} \emph{\text{``direction change,''}} or \emph{\text{``stop.''}} 
This structure is designed not only to enhance the model's ability to reason over instruction realization but also to diagnose predictions. The reason for the ground truth can be computed using the simulation's state.

\boldparagraph{{Timing Supervision}}
To efficiently issue instructions while facilitating temporal alignment, we introduce additional timing supervision-a binary classifier head determining whether an instruction should be issued. Specifically, we encode instructional cues as a 1D binary vector $\bi_t \in \{0,1\}^T$, where each entry indicates whether an instruction was given at a past timestep. This trace vector is encoded via an MLP and combined with visual encoder features to predict whether an instruction should be generated.
During inference, the timing prediction can also serve as a gating mechanism~\cite{sengupta2024unilcd} for reduced latency: if no instruction is needed according to the timing classifier, the model performs an early exit, bypassing the instruction generation stage and thereby reducing both inference time and computational overhead.

\boldparagraph{Loss Function} We supervise the model via Cross-Entropy (CE) loss over the generated output:

$$
\mathcal{L}_{\text{CE}} = \exp\left[ -\frac{1}{N_w} \sum_{n=1}^{N_w} \log P(w^n \mid w^1, \dotsc, w^{n-1}) \right]
\eqno{(1)}
$$
where $N_w$ is the total output length. We note that the generated output contains both reason supervision (which is not announced to the user) and output instruction. The overall loss is then:

$$
\mathcal{L}_{\mathrm{out}} =
\mathcal{L}^{\text{cot-reason}}_{\mathrm{CE}} +
\mathcal{L}^{\text{instruction}}_{\mathrm{CE}} +
\beta \mathcal{L}^{\text{issue}}_{\mathrm{BCE}}
\eqno{(2)}
$$

where we add the reason prediction and issuance classification (Binary Cross Entropy loss) as auxiliary tasks, and a constant hyperparameter $\beta$.

\section{Experiments}
\label{sec:experiments}
In this section, we first discuss our experimental setup, including the models and evaluation metrics. 
Next, we discuss our results with off-the-shelf and fine-tuned models in open and closed-loop settings.

\subsection{Experimental Setup}
\label{sec:setup}

\boldparagraph{Models}
We build on several state-of-the-art MLLM, such as ones take an image~\cite{zhou2024tinyllava,liu2024improved,gpt4o,zhou2024navgpt} or video~\cite{ren2024timechat,lin2024vila,gpt4o,zhang2024navid} as input.
We split our dataset into training (Town05 and Town10 in CARLA) and testing (Town02, Town03, and Town04). 
Models were initiated from an open-source checkpoint with four A6000 GPUs.
For MLLMs with an image input, we provide the current frame of the video.

\begin{table}[!t]
\captionsetup{width=\columnwidth}
\caption{\textbf{Open-Loop Assistive Instruction Generation Using Off-the-Shelf Models and Finetuned Models.}
We sequentially provide models with images and videos in the TIMELI benchmark to evaluate their safety-critical navigation performance.
The output of the off-the-shelf models is guided via in-context learning. 
While all models have access to goal coordinates, the finetuned model denoted by ``+'' incorporates instruction history, high-level plan, and reason, while those denoted by ``++'' additionally include timing supervision.}
\label{tab:result_00}
\resizebox{\columnwidth}{!}{%
\begin{tabular}{lcc|cccccc}
\toprule
  Model &
  Size &
  Modality &
  \multicolumn{1}{c}{BLEU-4} &
  \multicolumn{1}{c}{ROUGE-L} &
  \multicolumn{1}{c}{\begin{tabular}[c]{@{}c@{}}Timing\\F1\end{tabular}} &
  \multicolumn{1}{c}{\begin{tabular}[c]{@{}c@{}}Timing\\AUC\end{tabular}} & 
  \multicolumn{1}{c}{\begin{tabular}[c]{@{}c@{}}Conciseness\end{tabular}} \\
\midrule
\multicolumn{8}{c}{\textit{Off-the-Shelf Models}} \\
\midrule 
VILA~\cite{lin2024vila}            & 3B  & Image & 0.626 &  6.545 &  0.579 &  0.778 &  0.554 \\
TinyLLaVA~\cite{zhou2024tinyllava} & 3B  & Image & 0.000 &  5.728 & 0.361 & 0.500 &  0.264 \\
VILA~\cite{lin2024vila}            & 3B  & Video & 0.000 &  4.377 & 0.465 & 0.651 &  0.552 \\
LLaVA-v1.6~\cite{liu2024improved}  & 7B  & Image & 1.849 &  7.851 & 0.575 & 0.776 & 0.068 \\
NaVid~\cite{zhang2024navid}        & 7B  & Video & 0.000 &  6.659 & 0.384 & 0.606 & 0.046 \\
TimeChat~\cite{ren2024timechat}    & 7B  & Video & 0.000 &  7.423 & 0.379 & 0.536 & 0.026 \\
VILA~\cite{lin2024vila}            & 8B  & Image & 0.201 &  3.916 & 0.566 & 0. 788 & 0.010 \\
LLaVA-v1.6~\cite{liu2024improved}  & 13B & Image & 3.205 & 12.991 & 0.499 & 0.726 & 0.116 \\
NavGPT~\cite{zhou2024navgpt}       & -   & Image & 2.536 &  12.998 & 0.042 & 0.508 & 0.167 \\ 
GPT-4o~\cite{gpt4o}                & -   & Image & 7.395 &  20.059 & 0.451 & 0.654 & 0.220 \\
GPT-4o~\cite{gpt4o}                & -   & Video & 2.536 &  12.998 & 0.042 & 0.508 &  0.245 \\
\midrule
\multicolumn{8}{c}{\textit{Finetuned Models on TIMELI Benchmark}} \\
\midrule 
TinyLLaVA     & 3B  & Image & 0.000 &  0.000 & 0.000 & 0.500 & 0.000 \\
TinyLLaVA+    & 3B  & Image & 13.294 & 27.730 & 0.254 & 0.573 & \cellcolor{i2} 0.325 \\
VILA          & 3B  & Video & 8.618  & 20.063 & 0.540 & 0.726 &  0.048 \\
VILA+         & 3B  & Video & 10.916 & 23.789 & 0.410 & 0.629 & 0.300 \\
TimeChat      & 7B  & Video &  0.000 &   6.788 & 0.030 & 0.508 & 0.067 \\
TimeChat+     & 7B  & Video & 11.891 & 29.127 & 0.101 & 0.525 &  0.316  \\
LLaVA-v1.6    & 7B  & Image & 12.398 & 24.765 & 0.572 & 0.749 & 0.064 \\
LLaVA-v1.6+   & 7B  & Image & \cellcolor{i1} 20.390 & \cellcolor{i1} 38.250 & \cellcolor{i2} 0.776 & \cellcolor{i2} 0.869  & \cellcolor{i1} 0.395 \\
LLaVA-v1.6++  & 7B  & Image & \cellcolor{i2} 19.660 & \cellcolor{i2} 37.709 & \cellcolor{i1} 0.792 & \cellcolor{i1} 0.879 & 0.219 \\
\bottomrule
\end{tabular}%
}
\vspace{-0.6cm}
\end{table}
\boldparagraph{Metrics} 
We evaluate our model's performance based on timing, navigation instruction, and navigation performance.
For timing, we report the F1 score and AUC.
Timing metrics were calculated with a one-second timing tolerance, where predictions within one second of the ground truth are considered correct. 
We then employ standard language metrics for instruction evaluation, including BLEU-4~\cite{papineni2002bleu} and ROUGE-L~\cite{lin2004rouge}, computed only when the predicted timing falls within this tolerance as the model learns to omit instructions when not necessary.
We also report conciseness by identifying the number of overlapping words between the prediction and the ground truth after removing stopwords from both, then dividing this value by the prediction’s word count without stopwords. 
We leverage CARLA's interactive environment for closed-loop evaluation, measuring general VLN metrics~\cite{wang2022less}, including success rate, route completion, navigation score, collisions per minute, and instructions per minute to understand the models' guidance performance.
Finally, we define a Navigation Quality Score (NQS) metric:
$$
\text{NQS}
=
\left(
SR \cdot RC \cdot NS \cdot S_C \cdot S_I
\right)^{1/5}
\eqno{(3)}
$$
where $SR$, $RC$, and $NS$ denote success rate, route completion, and navigation score. The collision penalty and instruction efficiency are defined as
$$
S_C = \exp(-C/4), \quad
S_I = \exp\!\left(-\frac{(I-I^*)^2}{2\sigma^2}\right)
\eqno{(4)}
$$
where $C$ is collisions per minute, $I$ is instructions per minute, $I^*=20$ is the optimal instruction rate, and $\sigma=10$ controls the acceptable deviation.

\subsection{Results}
\boldparagraph{Off-the-Shelf Model Failure and Role of Domain Finetuning}
Table~\ref{tab:result_00} presents the comparison between off-the-shelf models and TIMELI fine-tuned models.
The off-the-shelf models were guided with an in-context learning method, which the prompt contains example inputs and outputs. 
We find current off-the-self models to perform poorly on our task, as indicated by the low BLEU-4 scores, and fail to provide instructions at the appropriate times, with a Timing F1 score near or below 0.50.
In our prompt and output ablations for the fine-tuned models, we observe a general trend where instruction history, high-level plan, and reason prediction output complement to improve model performance. 
 {Also, the model trained along with timing prediction increased their performance on the timing metric.}
Interestingly, image-based models outperform their video-based counterparts. 
We explain this due to the high task complexity and input dimensionality; we find video models to struggle in compressing long-term information and learning to attend to correct video elements. This finding highlights a future direction in more effective video modeling techniques beyond basic video descriptions.

\begin{table}[!t]
\caption{\textbf{Zero-Shot Transfer from Simulation to Real-World.} We evaluate models trained on the TIMELI benchmark in open-loop settings on real-world videos. We find models transfer knowledge from our extensive simulation benchmark to real-world scenarios without any further fine-tuning. 
}
\label{tab:result_youtube_validation}
\resizebox{\columnwidth}{!}{%
\begin{tabular}{lrc|ccccc}
\hline
  Model &
  Size &
  Modality &
  \multicolumn{1}{c}{BLEU-4} &
  \multicolumn{1}{c}{ROUGE-L} &
  \multicolumn{1}{c}{\begin{tabular}[c]{@{}c@{}}Timing\\F1\end{tabular}} &
  \multicolumn{1}{c}{\begin{tabular}[c]{@{}c@{}}Timing\\AUC\end{tabular}} &
  \multicolumn{1}{c}{\begin{tabular}[c]{@{}c@{}}Conciseness\end{tabular}} \\ 
  \hline

TinyLLaVA+   & 3B & Image &  0.065 & 1.592 & 0.454 & 0.500  & 0.019  \\ 
VILA+        & 3B & Video &  7.005 &  20.359 & 0.186 & 0.545  & 0.253 \\ 
TimeChat+    & 7B & Video & 0.001 & \cellcolor{i2} 22.995 & 0.025 & 0.506  & \cellcolor{i1} 0.383 \\
LLaVA-v1.6+  & 7B & Image & \cellcolor{i2} 8.478 & 22.587 & \cellcolor{i1} 0.732 & \cellcolor{i1} 0.815  & \cellcolor{i2} 0.278 \\
LLaVA-v1.6++ & 7B & Image & \cellcolor{i1} 9.479 & \cellcolor{i1} 23.970 & \cellcolor{i2} 0.672 & \cellcolor{i2} 0.771 & 0.266 \\
\hline
\vspace{-0.3cm}
\end{tabular}%
}

\end{table}

\boldparagraph{Real-World Transfer Evaluation} 
To evaluate sim-to-real transferability, we validated our models using annotated real-world videos (Table~\ref{tab:result_youtube_validation}). Interestingly, we find some models fine-tuned on purely synthetic data to transfer to real-world settings. 
While we observed performance degradation compared to the simulation environment, the degradation in image models was not severe, whereas video models showed a more significant drop in performance. 
We find that current video MLLMs inefficiently handle longer inputs (consistent with findings in other domains~\cite {liu2024lost}).

\begin{table}[!t]
\caption{
{\textbf{Closed-Loop Performance Using Finetuned Models.} We perform a study where pedestrians are controlled based on instructions generated from models to walk along 12 routes in the simulation environment (see metric definition in Sec.~\ref{sec:setup}). Note that some models may achieve low errors due to minimal progress along the route.}
}
\label{tab:result_02_closedloop}
\resizebox{\columnwidth}{!}{%
\begin{tabular}{lcc|c|ccccc}
\toprule
  Model &
  Size &
  Modality &
  NQS &
  \multicolumn{1}{c}{\begin{tabular}[c]{@{}c@{}}{Success}\\{Rate}\end{tabular}} &
  \multicolumn{1}{c}{\begin{tabular}[c]{@{}c@{}}{Route}\\{Completion}\end{tabular}} &
  \multicolumn{1}{c}{\begin{tabular}[c]{@{}c@{}}{Navigation}\\{Score}\end{tabular}} &
  Collision/Min &
  Instructions/Min 
  \\
\midrule
TinyLLaVA    & 3B  & Image & 0.00 & 0.00 & 0.01 & 0.01 & \cellcolor{i1} 0.00 & \cellcolor{i1} 0.00 \\
TinyLLaVA+   & 3B  & Image & \cellcolor{i2}0.81 & \cellcolor{i2} 0.47 & \cellcolor{i1} 0.93 & \cellcolor{i1} 0.92 &  0.51 & 18.91 \\
VILA         & 3B  & Video & 0.29 & 0.02 & 0.55 & 0.44  & 0.68 & 8.66 \\
VILA+        & 3B  & Video & 0.57 & 0.14 & 0.79 &  0.63 &  0.60 & 16.35 \\ 
TimeChat     & 7B  & Video  & 0.24 & 0.04 & 0.45 & 0.46 & \cellcolor{i2} 0.41 & \cellcolor{i2} 2.95 \\ 
TimeChat+    & 7B  & Video &0.38& 0.17 & 0.68 & 0.54 &  0.51 &  3.66 \\
LLaVA-v1.6   & 7B  & Image & 0.31& 0.04 & 0.36 & 0.29 &  0.54 & 12.65 \\
LLaVA-v1.6+  & 7B  & Image  & 0.79 & 0.44 & \cellcolor{i1} 0.93 & \cellcolor{i1} 0.92 &  0.69 & 19.57\\
LLaVA-v1.6++\ & 7B  & Image & \cellcolor{i1}0.82 & \cellcolor{i1} 0.50 & \cellcolor{i2} 0.92 & \cellcolor{i2} 0.91 & 0.60 & 21.22 \\
\bottomrule
\end{tabular}%
}
\vspace{-0.6cm}
\end{table}

\boldparagraph{Closed-Loop Evaluation in CARLA} 
Table~\ref{tab:result_02_closedloop} shows the result of the interactive closed-loop evaluation in 12 routes in the CARLA environment. 
We control the pedestrian by mapping the model-generated instructions to actions using a rule-based function that determines pedestrian movement.
To simulate realistic sensor measurements, we introduce Gaussian noise to the goal coordinates and pedestrian orientation. 
To simulate real-world uncertainty, we inject Gaussian localization noise (mean of 0\,m and SD of 2.0\,m) into the planner, and orientation noise (mean of 0$^{\circ}$ and SD of 5$^{\circ}$).
 {Our LLaVA-v1.6++ image-based model achieves a success rate of up to 50\% in guiding users to the goal without repetitive instructions.}
We also observe that success in language modeling metrics in open-loop does not always lead to better performance in closed-loop evaluation for some of the models, which is consistent with prior findings in embodied navigation~\cite{zhao2021evaluation}.

\vspace{-0.1cm}
\section{Conclusion}
\label{sec:conclusion}
An overarching goal in interactive vision-based systems is to develop adaptable models that do not require extensive fine-tuning or annotations across different settings. However, our findings reveal that current MLLMs struggle with fundamental concepts of assistive navigation, often failing to determine both when and what to instruct. These models tend to prioritize content generation while neglecting crucial temporal and contextual factors necessary for safe and effective assistance. While our model shows potential in supporting visually impaired individuals by automating navigation tasks, we acknowledge the inherent risks of AI-generated instructions. Misleading or poorly timed guidance could lead to hazardous situations, making reliability and safety paramount. We aim to enhance MLLM performance through improved reasoning and rigorous testing in diverse real-world conditions. We hope that our framework and tools will contribute to developing intelligent guidance systems capable of reasoning over diverse user needs, tasks, and environments. While we focus on the smartphone platform due to its scalability and low cost, future evaluations of generalization on other robotic platforms, such as ground robots~\cite{defazio2023seeing,wang2017enabling,zhang2020visual}, will be crucial to validate and broaden the applicability of our approach.

\boldparagraph{Acknowledgments}
We thank the National Science Foundation (IIS-2152077, IIS 2443719) for supporting this research.

\vspace{-0.1cm}
\bibliographystyle{IEEEtran}
\bibliography{main}




\end{document}